\documentclass[longbibliography,twocolumn,amsmath,amssym,amsthm,amsfonts,superscriptaddress]{revtex4-2}
\usepackage[colorlinks=true,urlcolor=blue,citecolor=blue,linkcolor=blue]{hyperref}
\usepackage{graphicx}
\usepackage{hyperref}
\usepackage{bm}

\DeclareMathOperator*{\argmax}{arg\,max}

\begin{document}
\title{Deep reinforcement learning for feedback control in a collective flashing ratchet}
\author{Dong-Kyum Kim}
\affiliation{Department of Physics, Korea Advanced Institute of Science and Technology, Daejeon 34141, Korea}
\author{Hawoong Jeong}
\email{hjeong@kaist.edu}
\affiliation{Department of Physics, Korea Advanced Institute of Science and Technology, Daejeon 34141, Korea}
\affiliation{Center for Complex Systems, Korea Advanced Institute of Science and Technology, Daejeon 34141, Korea}
\begin{abstract}
A collective flashing ratchet transports Brownian particles using a spatially periodic, asymmetric, and time-dependent on-off switchable potential. The net current of the particles in this system can be substantially increased by feedback control based on the particle positions. Several feedback policies for maximizing the current have been proposed, but optimal policies have not been found for a moderate number of particles. Here, we use deep reinforcement learning (RL) to find optimal policies, with results showing that policies built with a suitable neural network architecture outperform the previous policies. Moreover, even in a time-delayed feedback situation where the on-off switching of the potential is delayed, we demonstrate that the policies provided by deep RL provide higher currents than the previous strategies.
\end{abstract}

\maketitle
\newpage
\paragraph{Introduction.}
A flashing ratchet is a nonequilibrium model that induces a net current of Brownian particles in a spatially periodic asymmetric potential that can be temporally switched on and off~\cite{prost1994asymmetric,astumian1994fluctuation,astumian1997thermodynamics,tarlie1998optimal}. If one can access the position information of the particles, the current can be greatly improved by feedback control that switches the potential on-off based on the position information~\cite{cao2004feedback}. Feedback strategies for maximizing the current in flashing ratchets have been extensively studied~\cite{tarlie1998optimal,cao2004feedback,dinis2005closed,feito2006threshold,feito2009optimal,feito2007delay,craig2007effect,craig2008feedback,lopez2008real,roca2014optimal} due to the model's applicability in various disciplines~\cite{reimann2002brownian}; for instance, flashing ratchets have been used for explaining transport phenomena in biological processes such as ion pumping~\cite{siwy2002fabrication}, molecular transportation~\cite{kosztin2004fluctuation}, and by motor proteins~\cite{campas2006collective,brugues2009self,oriola2013cooperative,hwang2019structural}. However, the proposed feedback strategies~\cite{tarlie1998optimal,cao2004feedback,dinis2005closed,feito2006threshold,feito2009optimal,feito2007delay,craig2007effect,craig2008feedback} are not optimal policies for a moderate number of particles and require prior information of the system as well.

Thanks to the recent advances in deep learning~\cite{goodfellow2016deep}, physicists in diverse fields have been applying it to complex problems that are analytically intractable, e.g. glassy systems~\cite{bapst2020unveiling}, quantum matter~\cite{carrasquilla2020machine}, and others~\cite{carleo2019machine}. In particular, reinforcement learning (RL)~\cite{sutton2018reinforcement} has shown unprecedented success in previously unsolvable problems through combination with deep neural networks~\cite{mnih2015human,silver2016mastering,silver2018general,vinyals2019grandmaster}. This framework, so-called deep RL, has become a highly efficient tool for quantum feedback control, showing similar or better performance than previous handcrafted policies~\cite{fosel2018reinforcement,porotti2019coherent,niu2019universal,an2019deep,wang2019deep}. In this Letter, we employ deep RL to obtain optimal policies in the collective flashing ratchet model, and validate our approach by application to a time-delayed feedback situation that occurs in actual experiments~\cite{lopez2008real}.

\begin{figure}[!b]
\includegraphics[width=\columnwidth]{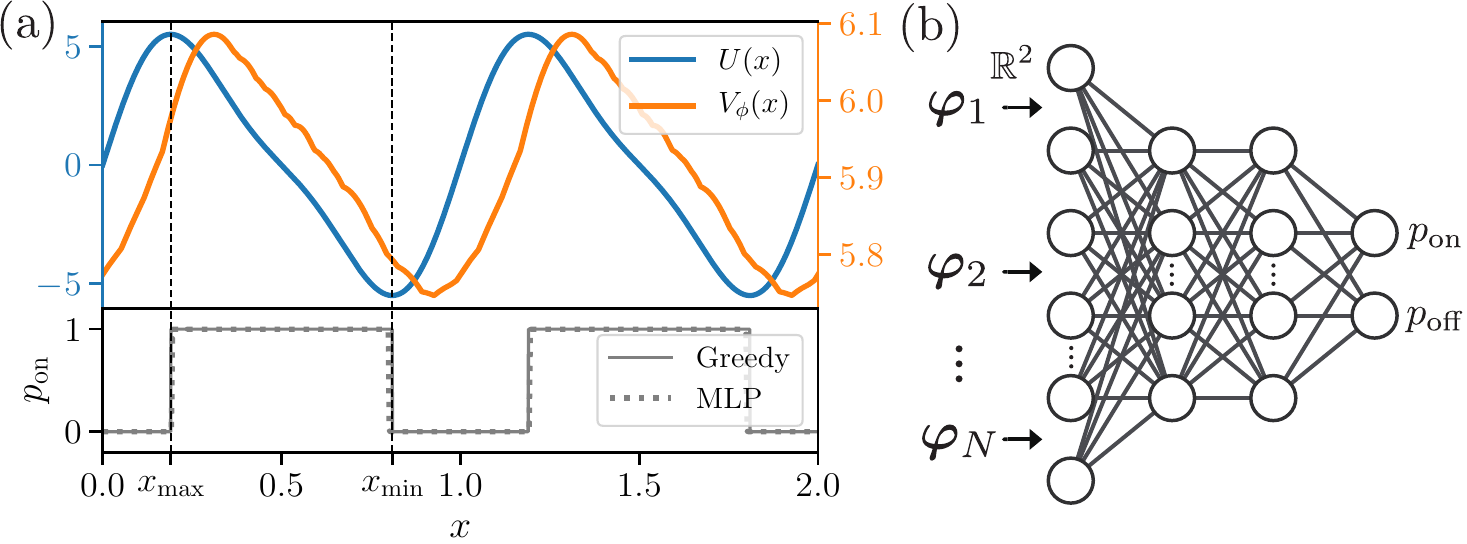}
\caption{(a) $N=1$ case. Top: Potential $U$ and trained value network $V_\phi$ as a function of position $x$ are denoted by blue and orange lines, respectively. Bottom: The solid line denotes the probability of switching on the potential ($p_{\rm on}$) as a function of $x$ for the greedy policy. The dotted line represents $p_{\rm on}$ of the trained MLP policy. (b) Illustration of a MLP with two hidden layers for the policy network $\pi_\theta$.}
\label{fig:fig1}
\end{figure}

\paragraph{Collective flashing ratchet.} 
We consider the collective flashing ratchet model~\cite{cao2004feedback}, which consists of an ensemble of $N$ non-interacting Brownian particles in contact with a heat bath at temperature $T$ and that drift in a spatially periodic asymmetric potential $U$. The dynamics of the $N$ particles is governed by the following overdamped Langevin equation:
\begin{equation}
\begin{gathered}
\eta \dot{x}_i(t) = \alpha(s_t) F(x_i(t)) + \xi_i(t); \\
s_t \equiv \{x_1(t), \dots, x_N(t)\}, \quad i=1, \dots, N,
\end{gathered}
\label{eq:eom}
\end{equation}
where $x_i(t)$ is the position of particle $i$, $\eta$ is the friction coefficient, and $\xi_i$ is a Gaussian noise with zero mean and correlation $\mathbb{E}[\xi_i(t)\xi_j(t')]=2\eta k_{\rm B}T \delta_{ij} \delta(t-t')$ where $\mathbb{E}$ denotes the ensemble average. Here, $\alpha$ is a deterministic control policy that depends on a set of positions $s_t$ with an output of $0$ (off) or $1$ (on). The force is given by $F(x)=-\partial_x U(x)$ with the potential [see Fig.~\ref{fig:fig1}(a)]
\begin{align}
    U(x) = U_0 \left[ \sin\left({\frac{2 \pi x}{L}}\right) + \frac{1}{4} \sin\left({\frac{4 \pi x}{L}}\right)  \right].
    \label{eq:potential}
\end{align}
In all simulations, we set $L=1$, $k_{\rm B}T = 1$, diffusion coefficient $D=k_{\rm B}T/\eta=1$, $U_0=5k_{\rm B}T$, and time step size $\Delta t=10^{-3} L^2/D$. The current of the particles in steady state under policy $\alpha$ is denoted as
\begin{align}
    \mathbb{E}_\alpha[\dot{x}] \equiv \mathbb{E}_\alpha\left[\frac{1}{N}\sum_{i=1}^{N} \dot{x}_i\right] \quad ({\rm Unit}: D/L).
    \label{eq:current}
\end{align}

Various policies for maximizing the current~\eqref{eq:current} have been proposed as follows: the periodic switching policy~\cite{tarlie1998optimal}, maximizing instantaneous current (greedy policy)~\cite{cao2004feedback}, threshold policy~\cite{dinis2005closed,feito2006threshold,feito2009optimal}, and Bellman’s criterion~\cite{roca2014optimal}.

The periodic switching policy~\cite{tarlie1998optimal} is $\alpha(t)=1$ for $t \in [0, \mathcal{T}_{\rm on})$, $\alpha(t)=0$ for $t \in [\mathcal{T}_{\rm on}, \mathcal{T}_{\rm on} + \mathcal{T}_{\rm off})$, and periodic $\alpha(t+\mathcal{T}_{\rm on}+\mathcal{T}_{\rm off})=\alpha(t)$ with optimal periods $\mathcal{T}_{\rm on} \approx 0.03L^2/D$ and $\mathcal{T}_{\rm off} \approx 0.04 L^2/D$. For any $N$, this policy gives the current $\mathbb{E}_\alpha[\dot{x}] \approx 0.862 D/L$ because it does not depend on the position but only time.

The greedy policy~\cite{cao2004feedback} is defined as $\alpha(s_t)=\Theta(f(s_t))$, where $f(s_t)=\sum_{i=1}^N F(x_i(t))/N$ is the mean force and $\Theta$ is the Heaviside function given by $\Theta(z)=1$ if $z>0$ or else 0. While the greedy policy is the optimal one for $N=1$, this policy is outperformed by the periodic switching policy for large $N$. 

The threshold policy~\cite{dinis2005closed,feito2006threshold,feito2009optimal} is $\alpha(s_t)=0$ if $f(s_t) \le u_{\rm on}$ when $f(t)$ is decreasing, and $\alpha(s_t)=1$ if $f(s_t) \ge u_{\rm off}$ when $f(t)$ is increasing, with thresholds $u_{\rm on} \ge 0$ and $u_{\rm off} \le 0$. The threshold policy with optimal thresholds gives mostly similar performance to the greedy policy for $N < 10^2$--$10^3$ and is better than the greedy policy for larger $N$. It is also optimal for $N = \infty$, which is equivalent to the periodic switching policy.

Neither greedy nor threshold policy is optimal for finite $N > 1$. Roca {\it et al.}~\cite{roca2014optimal} proposed a general framework for finding the optimal policy via Bellman's principle, and found it for $N=2$ using numerical integration. However, this numerical method requires prior information of the model and is computationally infeasible for large $N$ due to the curse of dimensionality.

\paragraph{Methods.}
We employ the actor-critic algorithm, which is one of the policy gradient methods in RL~\cite{sutton2018reinforcement}, together with deep neural networks to find the optimal policies in the collective flashing ratchet for any $N$.

To formulate this problem in RL language, we define the reward as the total mean displacement of the particles: 
\begin{align}
    r_t = \frac{1}{N}\sum_{i=1}^N \left(x_i(t+\Delta t) - x_i(t)\right).
    \label{eq:reward}
\end{align}
The total discounted reward from time $t$, called return, is $G_t = \sum_{k=0}^{\infty}\gamma^k r_{t+(k+1)\Delta t}$ where $\gamma \in [0, 1)$ is the discounting factor and we set $\gamma=0.999$. We build a policy network $\pi_{\theta}$, called actor, where $\theta$ denotes the trainable neural network parameters, that takes system state $s$ as an input. The outputs $\pi_{\theta}(s)=(p_{\rm on}, p_{\rm off})$ are the probabilities for switching the potential on or off [see Fig.~\ref{fig:fig1}(b)]. We sample the on-off probability from $\pi_\theta(s_t)$ every $t$ in the training process. 

The goal in RL is obtaining the optimal policy $\pi^{*}$ that maximizes the expected total future reward, i.e. $\pi^{*}=\argmax_{\pi} \mathbb{E}_{\pi}[G_t]$. If the equation of motion is known, $\mathbb{E}_{\pi}[G_t]$ can be numerically calculated using Bellman’s equation~\cite{roca2014optimal}. However, in this work, we assume that we can only access the system state $s_t$ and reward $r_t$. In such case, called model-free RL, we need an estimator $V_{\phi}$ for a {\it value function}:
\begin{align}
    V^{\pi}(s_t) = \mathbb{E}_{\pi}[G_t|s_t],
    \label{eq:value-ftn}
\end{align}
which is the expected return given state $s_t$ under a policy $\pi$. The estimator $V_{\phi}$, called value network or critic, where $\phi$ denotes the trainable parameters, is also built with another neural network.

There are various optimization methods for the actor-critic algorithm~\cite{SpinningUp2018}. Among them, we employ proximal policy optimization~\cite{schulman2017proximal}, which is widely used in RL because of its scalability, data efficiency, and robustness for hyperparameters (see Supplemental Material~\cite{SM} for training details). After the training process is complete, we test the policy deterministically, i.e.
\begin{align}
\alpha(s_t) = \left\{ 
\begin{array}{ll} 
    1 & {\rm if}~p_{\rm on} > 0.5  \\ 
    0 & {\rm if}~p_{\rm off} > 0.5, 
\end{array} \right.
{\rm where}~ (p_{\rm on}, p_{\rm off}) = \pi_{\theta}(s_t). \nonumber
\end{align}

\begin{figure}[!t]
\includegraphics[width=\columnwidth]{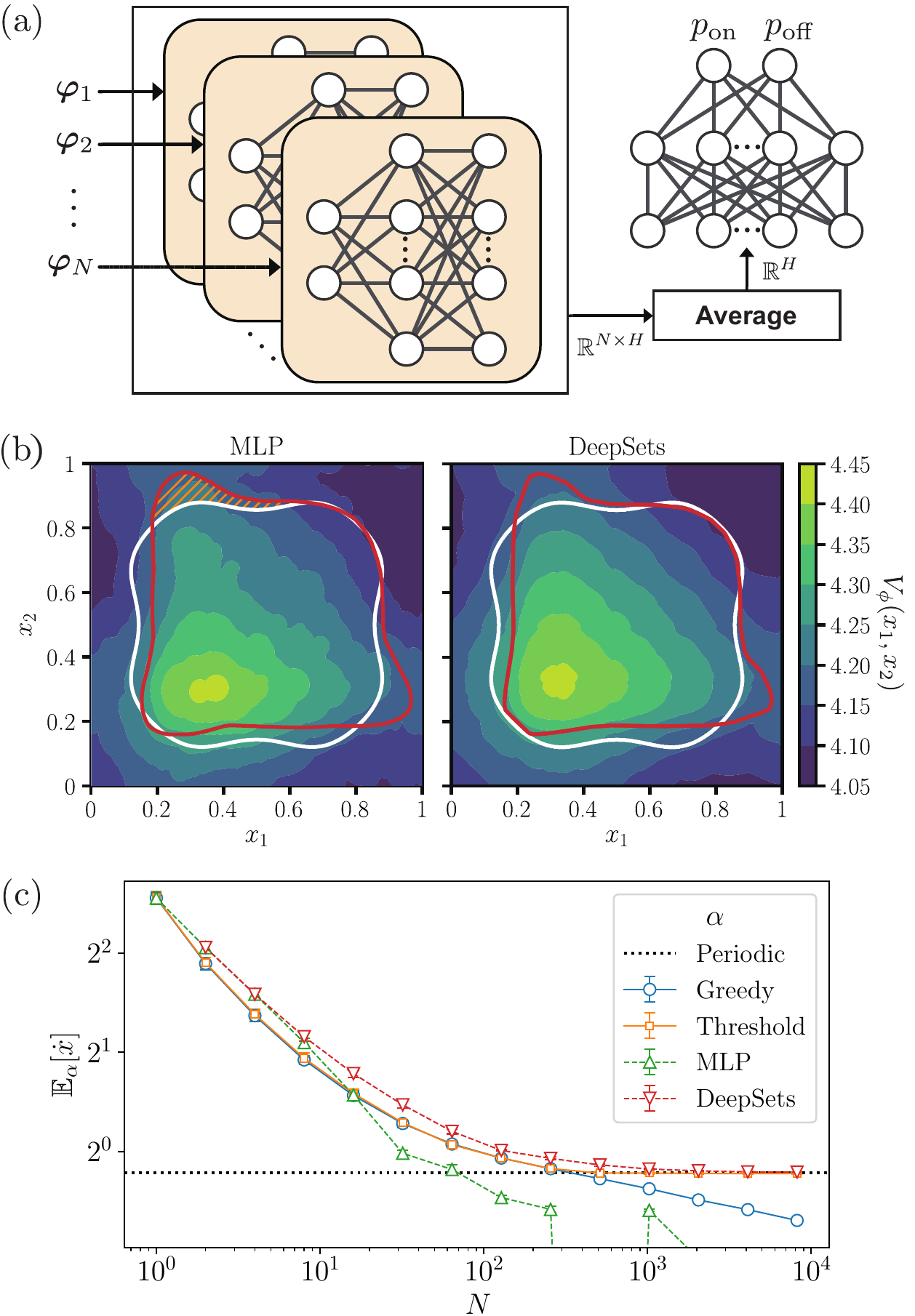}
\caption{(a) DeepSets architecture for the policy network $\pi_\theta$. $H$ is the number of hidden units for each layer. (b) Decision boundaries from a trained MLP (left) and trained DeepSets (right) for $N=2$. The white contour denotes where the mean force $f(x_1, x_2)$ is zero. The red contour is $p_{\rm on}=0.5$ from the trained policy network $\pi_\theta$. The color gradient represents the trained value network $V_\phi$. (c) Current $\mathbb{E}_{\alpha}[\dot{x}]$ as a function of $N$ for each policy $\alpha$. Throughout this work, error bars represent the standard deviation of the current measured from the realized trajectory ensemble over the period $t = 50L^2/D$.}
\label{fig:deepsets}
\end{figure}

\paragraph{Neural network architecture.}
First, we employ multilayer perceptron (MLP) architecture for the policy network $\pi_{\theta}$ and value network $V_{\phi}$ [see Fig.~\ref{fig:fig1}(b)]. The configuration details of the neural network architectures are given in the Supplemental Material~\cite{SM}. Using the periodicity of the potential $U(x)$, we transform the state $s_t$ into the input feature $\psi_t = [\bm{\varphi}_1(t), \bm{\varphi}_2(t), \dots, \bm{\varphi}_N(t)]$ for neural network input where
\begin{align}
    \bm{\varphi}_i(t) = \left[\cos\left(\frac{2 \pi x_i(t)}{L}\right), \sin\left(\frac{2 \pi x_i(t)}{L}\right) \right].
\end{align}
Therefore, the input dimension of the MLP is $2N$ and the output dimension is two for $\pi_{\theta}$. The value network $V_\phi$ has the same configuration except for having an output dimension of one rather than two. We note that the discounting factor $\gamma=0.999$, which indicates the return $G_t$, can effectively be considered as the total mean displacement between $t$ and $t + \Delta t/(1-\gamma)$. Accordingly, $V_{\phi}(\psi_t)$ can be interpreted as the expected current given $\psi_t$ because the time step size is $\Delta t=10^{-3} L^2/D$. 

For the $N=1$ case, Fig.~\ref{fig:fig1}(a) shows that the trained $\pi_\theta$ agrees with the greedy policy (bottom panel), while $V_{\phi}$ is slightly shifted to the right from potential $U$ (top panel). This is because, at the top of the potential valley ($x_{\rm max}$), the particle can slide to the right or left with a 50/50 chance, and therefore the expected current is maximum slightly right of $x_{\rm max}$.

For the $N=2$ case, as shown in the left panel of Fig.~\ref{fig:deepsets}(b), the greedy policy switches on (off) the potential when the particles are inside (outside) the white contour. On the other hand, the decision boundary of the trained MLP policy $\pi_{\theta}$ (red contour) agrees with the policy discovered by Roca {\it et al.}~\cite{roca2014optimal} and shows better performance than the greedy policy by considering the future expected current. For instance, in the orange dashed area, the instantaneous net current will be negative because the mean force $f(x_1, x_2)$ is negative when the potential is on. But considering each particle with a long-term view, particle 1 and particle 2 are located on the downhill of the potential ($x_{\rm max} < x < x_{\rm min}$) and near the minimum ($x_{\rm min}$), respectively; while particle 2 will soon reach $x_{\rm min}$ and become trapped in the potential well, particle 1 can keep moving down along the potential~\cite{roca2014optimal}.

However, the decision boundary (red contour) and $V_\phi$ (color gradient) are not symmetric over the line $x_1=x_2$ [see Fig.~\ref{fig:deepsets}(b), left] because MLP outputs are not permutation invariant to the order of the elements in the input feature $\psi_t$. To address this issue, we employ a permutation invariant architecture, called DeepSets~\cite{zaheer2017deepset}, for the policy and value networks. In this architecture [see Fig.~\ref{fig:deepsets}(a)], each element $\varphi_i$ in the input feature $\psi_t$ is independently fed into a single MLP (beige), and the outputs of the MLP are averaged over the elements and then fed to an another MLP. By using DeepSets for training, the decision boundary and $V_{\phi}$ show perfect symmetry over the $x_1=x_2$ line [see Fig.~\ref{fig:deepsets}(b), right].

Now we apply these methods for $N=2^2$, $2^3$, ... $2^{13}$, and compare the training results with the greedy (blue circles), threshold (orange squares), and periodic switching (black dotted line) policies in Fig.~\ref{fig:deepsets}(c). Results show that the trained MLP policies (green triangles) outperform the greedy and threshold policies for $N < 10$, but perform poorly for $N > 10$ due to the lack of permutation invariance. On the other hand, the trained DeepSets policies (red triangles) outperform the other policies for any $N>1$ while converging to the periodic policy as $N$ increases (see Fig. S1, Supplemental Material~\cite{SM}). We have also verified that deep RL works well for the sawtooth potential (Fig. S2, Supplemental Material~\cite{SM}).

\paragraph{Time-delayed feedback.}
In an actual experiment, there is an inevitable time-delay between the measurement and the feedback due to the calculation time in the feedback algorithm~\cite{feito2007delay,craig2007effect,craig2008feedback,lopez2008real}. To verify that deep RL is applicable to such a realistic situation, we consider a feedback time-delay $\tau$ in Eq.~\eqref{eq:eom}, i.e. $\alpha(s_t)$ is replaced by $\alpha(s_{t-\tau})$. In this case, the maximal net displacement (MND) policy~\cite{craig2008feedback}, defined by
\begin{align}
    \alpha(s_{t-\tau})=\Theta\left(\sum_{i=1}^{N} d(x_i(t-\tau))\right),
    \label{eq:MND}
\end{align}
where the displacement function is $d(x) = x_{\rm min} + x_0 - x$ for $x_{\rm max} < x \leq x_{\rm max}+L$ and periodic $d(x)=d(x+L)$, can perform better than the greedy policy for $\tau > 0$ with optimal $x_0 < 0$~\cite{lopez2008real}. This can be considered as a $\tau$-delayed greedy policy because it predicts the arrival of the particles at $x_{\rm min}$ after $\tau$ from $x_0+x_{\rm min}$. We train the neural networks for $N=1, 2^1, 2^2, ..., 2^5$ with time-delay $\tau$ in the range of $0.00$--$0.05L^2/D$, and compare them with the greedy policy and the MND policy with optimal $x_0$.

\begin{figure}[!t]
\includegraphics[width=\columnwidth]{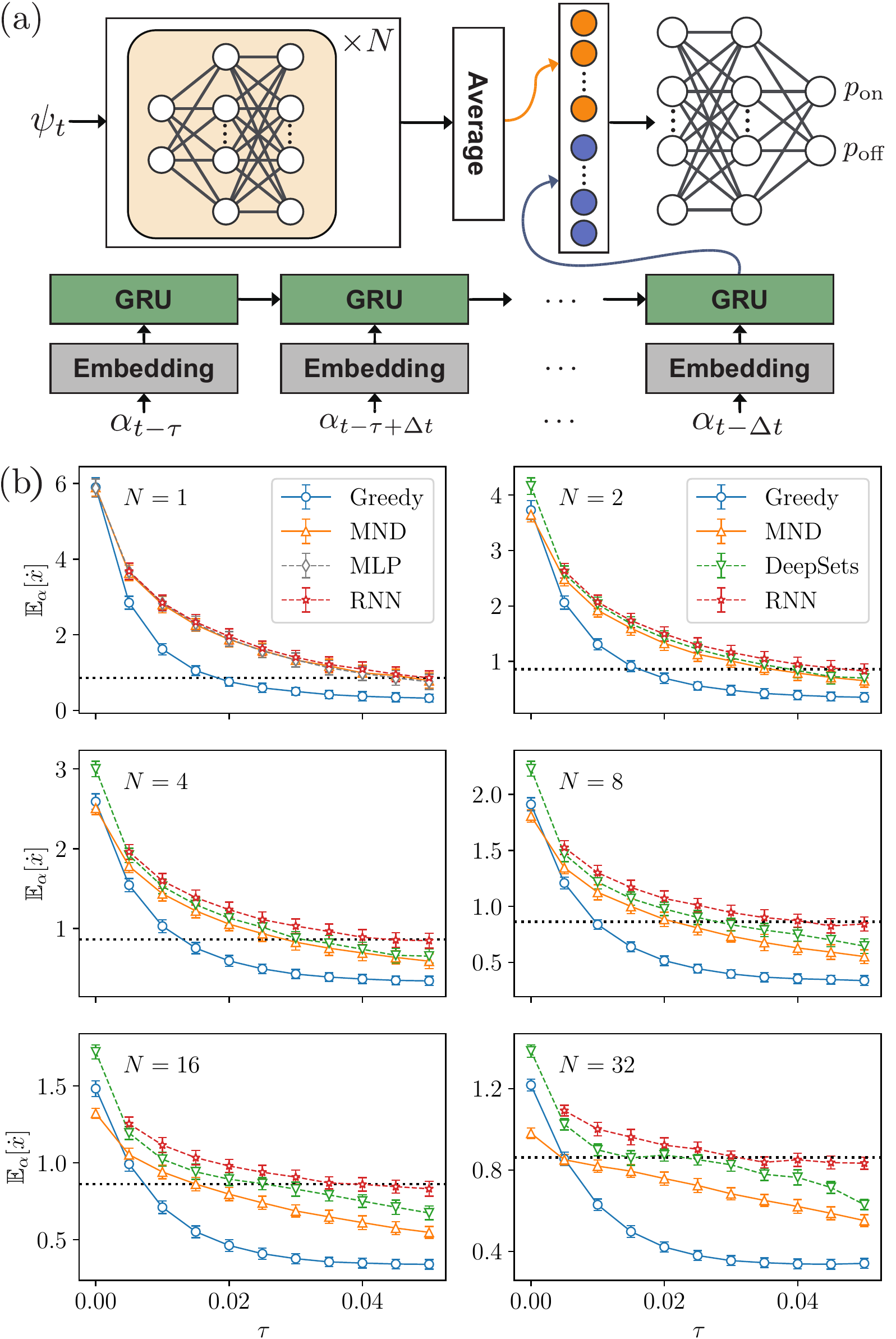}
\caption{(a) Architecture of policy network $\pi_{\theta}$ augmented with an RNN. (b) Time-delayed feedback results for the greedy, MND, MLP (only for $N=1$), DeepSets (for $N>1$), and RNN policies at increasing $N$. The black dotted lines denote the current of the periodic switching policy. }
\label{fig:time-delayed}
\end{figure}

For the time-delayed $N=1$ case [see Fig.~\ref{fig:time-delayed}(b), first row], the results show that the trained MLP policies (gray diamonds) agree with the MND policy (orange triangles) and perform better than the greedy policy (blue circles). For $N=2$, the trained DeepSets policies (green triangles) outperform the greedy policy and are slightly better than the MND policy.

While the actor-critic algorithm assumes that the feedback-controlled system is a Markov decision process (MDP), the delayed-feedback process is not a MDP because the next state $s_{t+\Delta t}$ not only depends on the previous state $s_t$ but also the history of the on-off information. This problem can be reformulated as a MDP by augmenting the input feature $\psi_t$ with the on-off history~\cite{katsikopoulos2003markov}. Here, the $d$-step augmented state at time $t$ is defined as 
\begin{align}
    I_t = (\alpha_{t-\tau}, \alpha_{t-\tau+\Delta t}, \dots, \alpha_{t-\tau + (d-1)\Delta t}, \psi_t), \quad d=\tau/\Delta t. \nonumber
    \label{augmented_state}
\end{align}

In order to efficiently handle the augmented state, we build the policy network with a recurrent neural network (RNN). We employ an embedding layer to transform the discrete variable $\alpha$ into a continuous variable, and we use a gated recurrent unit (GRU)~\cite{cho2014learning}, a widely used gating mechanism in RNNs due to its parameter efficiency and good performance on the sequential datasets, for the RNN. As shown in Fig.~\ref{fig:time-delayed}(a), we concatenate the output vectors from DeepSets (orange nodes) and the RNN (blue nodes), where DeepSets and the RNN encode the position information $\psi_t$ and potential on-off history, respectively. We then feed the concatenated vector to a MLP. See the Supplemental Material~\cite{SM} for the configuration details. As can be seen in Fig.~\ref{fig:time-delayed}(b), the trained RNN policies (red stars) show slightly better performance than the other policies for $N=1$ and noticeably better performance than the others for $N=2$. And also, the RNN policies outperform the greedy, MND, and DeepSets policies for the $N=4,8,16,32$ cases.

\paragraph{Conclusions and outlook.}
We have tackled the problem of finding an improved policy for maximizing the current in the collective flashing ratchet model through deep RL. Unlike the previous model-based method~\cite{roca2014optimal}, the model-free RL approach used in this study does not require information on the parameters of the system (e.g. potential, diffusion coefficient, and others). The deep RL approach makes it is possible to find state-of-the-art feedback strategies using suitable neural network architectures through training only in the process of interacting with the environment. Also, we have demonstrated that deep RL outperforms the previous strategies in a time-delayed feedback situation; therefore, we expect that this study can be effectively applied experimentally.

Although feedback control in the collective flashing ratchet can induce an effective coupling between non-interacting particles, molecular motors like kinesin, for example, explicitly interact with each other via hard-core repulsion. According to previous studies on interacting molecular motors~\cite{campas2006collective,brugues2009self,oriola2013cooperative}, their cooperative behavior can enhance transportation ability several times or more compared to individual motors. Further research applying deep RL on interacting molecular motors will be intriguing.

Another interesting future task would be the application of deep RL to a collective flashing ratchet in which a time-periodic external driving force acts on the particles~\cite{feito2009rocking}. A ratchet-like mechanism for transportation in the cell membrane (such as ion pumping~\cite{siwy2002fabrication} or glycerol transportation~\cite{kosztin2004fluctuation}) can improve the current via the periodic driving force. Therefore, investigating whether a deep RL agent can exploit not only fluctuations in the environment but also time-dependent environmental dynamics is expected to aid the understanding of such biological processes.

In real-world scenarios, there may be measurement or feedback errors due to instrument noise~\cite{cao2009information,cao2009thermodynamics,sagawa2012nonequilibrium}. Such cases are not only important in physics, e.g. information thermodynamics~\cite{parrondo2015thermodynamics}, but also in RL for real-world applications~\cite{dulacarnold2020empirical}. Therefore, it will also be an interesting future work to study RL from a thermodynamics perspective; we expect that the collective flashing ratchet model can be utilized as a useful environment to benchmark RL algorithms in such situations.

The results of all runs and the code implemented in \texttt{PyTorch}~\cite{pytorch2019} are available in Ref.~\cite{kim2020repo}.

\paragraph{Acknowledgments.}
This study was supported by the Basic Science Research Program through the National Research Foundation of Korea (NRF) (Grant No. NRF-2017R1A2B3006930).

\end{document}


\title{Supplemental Material: Deep reinforcement learning for feedback control in a collective flashing ratchet}
\author{Dong-Kyum Kim}
\affiliation{Department of Physics, Korea Advanced Institute of Science and Technology, Daejeon 34141, Korea}
\author{Hawoong Jeong}
\email{hjeong@kaist.edu}
\affiliation{Department of Physics, Korea Advanced Institute of Science and Technology, Daejeon 34141, Korea}
\affiliation{Center for Complex Systems, Korea Advanced Institute of Science and Technology, Daejeon 34141, Korea}
\maketitle

\setcounter{equation}{0}
\setcounter{figure}{0}
\setcounter{table}{0}
\setcounter{page}{1}

\renewcommand{\theequation}{S\arabic{equation}}
\renewcommand{\thefigure}{S\arabic{figure}}

\section{Training details}
We use the proximal policy optimization (PPO)~\cite{schulman2017proximal} algorithm implemented in Ref.~\cite{SpinningUp2018}. The PPO algorithm updates the parameter $\theta_i$ of policy network $\pi_{\theta}$ at epoch $i$ by the following equations: 
\begin{align}
    \theta_{i+1} &= \argmax_{\theta} L(\theta), \\
    L(\theta) &= \underset{(s,a) \sim \pi_{\theta_i}}{\mathbb{E}} \left[ \min\left( \frac{\pi_{\theta}(s, a)}{\pi_{\theta_i}(s, a)}  A^{\pi_{\theta_i}}(s,a),~g(\epsilon, A^{\pi_{\theta_i}}(s,a)) \right) \right], \label{eq:L-clip}
\end{align}
where $\pi_\theta(s, a)$ is the probability of choosing action $a$ given the state $s$, and $\epsilon$ is a hyperparameter that restricts how much the new policy can be changed from the old policy. Function $g$ is defined as 
\begin{align}
    g(\epsilon, A) = \left\{ \begin{array}{ll} (1 + \epsilon) A & ~{\rm if}~A \geq 0 \\ (1 - \epsilon) A & ~{\rm if}~A < 0. \end{array} \right.
\end{align}
Here, $A^{\pi}(s, a)$ is the advantage function that indicates how much better or worse the action $a$ is than the other actions on average for the present policy $\pi$ and given state $s$. In this section, $t$ represents the time step, rather than time, for convenience. The definition of $A^{\pi}(s_t, a_t)$ is
\begin{align}
    A^{\pi}(s_t, a_t) = Q^{\pi}(s_t, a_t) - V^{\pi}(s_t),
\end{align}
where $Q^{\pi}(s_t, a_t) = \mathbb{E}_{\pi}[G_t|s_t, a_t]$ and $V^{\pi}(s_t) = \mathbb{E}_{\pi}[G_t|s_t]$ are the action-value function and the value function, respectively. The return $G_t$ is defined as $G_t = \sum_{l=0}^{\infty}\gamma^l r_{t+l+1}$ where $\gamma$ is the discounting factor. PPO uses the generalized advantage estimator (GAE)~\cite{schulman2016gae} $\hat{A}_t^{{\rm GAE}(\gamma, \lambda)}$ for an accurate evaluation of the advantage function. $\hat{A}_t^{{\rm GAE}(\gamma, \lambda)}$ is defined as
\begin{align}
    \hat{A}_t^{{\rm GAE}(\gamma, \lambda)} &= \sum_{l=0}^{\infty} (\gamma \lambda)^l \delta_{t+l}^{V}, \label{eq:gae} \\
    \delta_t^{V} &= \underbrace{r_t + \gamma V^{\pi}(s_{t+1})}_{=Q^{\pi}(s_t, a_t)} - V^{\pi}(s_t), \label{eq:delta}
\end{align}
where $\lambda \in (0,1]$ is a hyperparameter for adjusting the bias-variance tradeoff in $\hat{A}_t^{{\rm GAE}(\gamma, \lambda)}$. See Algorithm~\ref{alg1} for the detailed training procedure. The policy and value networks are initialized with the default random initialization setup in \texttt{PyTorch}~\cite{pytorch2019}. Two Adam~\cite{kingma2015adam} optimizers are used respectively for the policy and value networks training. During the stochastic gradient ascent stage for the policy network (step 4 in Algorithm~\ref{alg1}), the early stopping method is applied to prevent the updated policy from going too far from the old policy~\cite{SpinningUp2018}. Early stopping means that if the Kullback--Leibler divergence (KLD) from the updated policy $\pi_{\theta}$ to the old policy $\pi_{\theta_i}$
\begin{align}
    D_{\rm KL}(\pi_{\theta_i}\parallel \pi_{\theta}) = \underset{(s, a) \sim \pi_{\theta_i}}{\mathbb{E}} \left[ \log{\frac{\pi_{\theta_i}(s, a)}{\pi_{\theta}(s, a)}} \right]
\end{align}
exceeds the threshold $1.5\times d_{\rm targ}$, then the gradient steps stop. See Table~\ref{table:parameters} for the values of all the hyperparameters used in the training. For each $N$, we run Algorithm~\ref{alg1} five times independently with five different random seeds, and we pick the policy and value network that show the best performance among the five. All reported results in this paper are from the best-performing neural networks. Each run was conducted on a single NVIDIA TITAN V GPU.

\begin{algorithm}[H]
    \caption{Proximal policy optimization algorithm}
    \label{alg1}
\begin{algorithmic}[1]
    \REQUIRE Environment, policy network $\pi_{\theta}$, value network $V_{\phi}$, optimizer for $\theta$, optimizer for $\phi$
    \FOR{$i = 1,2, \dots, \mathcal{E}$} 
    \STATE Run simulation of multiple trajectories simultaneously under the present policy network $\pi_{\theta_i}$ in the environment and collect the dataset ${\mathcal D}_i = \{\Gamma^{(m)}\}_{m=1}^{m=M}$ where $M$ is the number of trajectories. Here, $\Gamma$ denotes the trajectory of the state, action, and reward, and can be represented as $\Gamma=[s_1, a_1, r_1, s_2, a_2, r_2, s_3, \dots, r_{\rm T}, s_{{\rm T}+1}]$ where ${\rm T}$ is the length of a single trajectory.
    \STATE Compute estimated return 
    \begin{equation*}
        \hat{G}_t=\sum_{l=0}^{{\rm T}-t-1}\left(\gamma^l r_{t+l+1}\right) + \gamma^{{\rm T}-t} V^{\pi}(s_{{\rm T}+1}),
    \end{equation*} 
    $\delta_t^{V}$~\eqref{eq:delta}, and $\hat{A}_t^{{\rm GAE}(\gamma, \lambda)}$~\eqref{eq:gae} for all time step $t$ and trajectory $\Gamma$ in ${\mathcal D}_i$ using the present value network $V^{\pi}=V_{\phi_i}$.
    \STATE Update the policy network by maximizing the estimated $L(\theta)$~\eqref{eq:L-clip}:
        \begin{equation*}
        \theta_{i+1} = \argmax_{\theta} \frac{1}{M {\rm T}} \sum_{\Gamma \in {\mathcal D}_i} \sum_{t=1}^{\rm T} \min\left(
            \frac{\pi_{\theta}(s_t, a_t)}{\pi_{\theta_i}(s_t, a_t)}  \hat{A}_t^{{\rm GAE}(\gamma, \lambda)},~ g(\epsilon, \hat{A}_t^{{\rm GAE}(\gamma, \lambda)})
        \right),
        \end{equation*}
        via the optimizer for $\theta$ with a mini-batch size of $\mathcal{B}$.
    \STATE Update the value network by minimizing the mean-squared error:
        \begin{equation*}
        \phi_{i+1} = \argmin_{\phi} \frac{1}{M {\rm T}} \sum_{\Gamma \in {\mathcal D}_i} \sum_{t=1}^{\rm T}\left( \hat{G}_t - V_{\phi} (s_t) \right)^2,
        \end{equation*}
        via the optimizer for $\phi$ with a mini-batch size of $\mathcal{B}$.
    \ENDFOR
\end{algorithmic}
\end{algorithm}

\begin{table}[H]
\centering
\begin{tabular}{ccc}
\begin{tabular}{l|l} 
\hline
Hyperparameter & Value \\
\hline
Trajectory length ${\rm T}$ & 2000 \\
Number of epochs $\mathcal{E}$ & 400 \\
Discounting factor $\gamma$ & 0.999 \\
GAE parameter $\lambda$ & 0.95 \\
Clipping parameter $\epsilon$ & 0.2 \\
Target KLD for early stopping $d_{\rm targ}$ & 0.01 \\
Training iterations for $\pi_{\theta}$ per epoch & 625 \\
Training iterations for $V_{\phi}$ per epoch & 625 \\
Learning rate for $\pi_{\theta}$ & $3 \times 10^{-4}$ \\
Learning rate for $V_{\phi}$ & $10^{-3}$ \\
\hline
\end{tabular} &
\begin{tabular}{l|l|l} 
\hline
$N$ & $M$ & $\mathcal{B}$ \\
\hline
1 & 1024 & 4096 \\
2 & 512 & 4096 \\
4 & 256 & 4096 \\
8 & 128 & 4096 \\
16 & 64 & 2048 \\
32 & 32 & 1024 \\
64 & 16 & 512 \\
\hline
\end{tabular} &
\begin{tabular}{l|l|l} 
\hline
$N$ & $M$ & $\mathcal{B}$ \\
\hline
128 & 8 & 256 \\
256 & 8 & 256 \\
512 & 8 & 256 \\
1024 & 8 & 256 \\
2048 & 8 & 256 \\
4096 & 8 & 256 \\
8192 & 8 & 256 \\
\hline
\end{tabular} \\
\end{tabular}
\caption{Left: Hyperparameters. The hyperparameters not listed in this table are set as defaults in \texttt{PyTorch}. Right: The number of trajectories $M$ and mini-batch size $\mathcal{B}$ for each number of particles $N$.}
\label{table:parameters}
\end{table}

\section{Architecture configurations}
We use ReLU~\cite{nair2010rectified} as the activation function for the policy network $\pi_{\theta}$ and value network $V_{\phi}$. See Tables~\ref{table:mlp},~\ref{table:deepsets}, and~\ref{table:RNN} for configuration details of the MLP, DeepSets, and RNN policy networks, respectively. The policy network $\pi_{\theta}$ computes the on-off probabilities using the softmax function in the output layer. The value network $V_{\phi}$ has the same configuration except for having an output dimension of one rather than two. We set the number of hidden units to $H=64$ and the embedding dimension to $E=16$. Here, $\bm{\alpha}_t^d$ is the potential on-off history:
\begin{align}
    \bm{\alpha}_t^d = (\alpha_{t-\tau}, \alpha_{t-\tau+\Delta t}, \dots, \alpha_{t-\tau + (d-1)\Delta t}), \quad d=\tau/\Delta t.
\end{align}
\begin{table}[H]
\centering
\begin{tabular}{l|l|l}
\hline
\multicolumn{3}{c}{
\centering
\textbf{MLP policy network}} \\
\hline
    Layer & \makecell{Output dim} & Activation function\\
\hline
Input $\psi_t$ & $2N$ &   \\
Fully-connected & $H$ & ReLU\\
Fully-connected & $H$ & ReLU\\
Output layer & $2$ &  None\\
\hline
\end{tabular}
\caption{Two-hidden-layer MLP configuration.}
\label{table:mlp}
\end{table}

\begin{table}[H]
\centering
\begin{tabular}{l|l|l}
\hline
\multicolumn{3}{c}{
\centering
\textbf{DeepSets policy network}} \\
\hline
    Layer & \makecell{Output dim} & Activation function\\
\hline
Input $\psi_t$ & $N \times 2$ &   \\
Fully-connected & $N \times H$ & ReLU\\
Fully-connected & $N \times H$ & None\\
Average & $H$ & \\
Fully-connected & $H$ & ReLU\\
Output layer & $2$ &  None\\
\hline
\end{tabular}
\caption{DeepSets configuration.}
\label{table:deepsets}
\end{table}

\begin{table}[H]
\centering
\begin{tabular}{c|l|l|l}
\hline
\multicolumn{4}{c}{
\centering
\textbf{RNN policy network}} \\
\hline
    \makecell{Module} & \makecell{Layer} & \makecell{Output dim} & Activation function\\
\hline
\multirow{4}{*}{$\rm{DeepSets}(\psi_t)$} & Input $\psi_t$ & $N \times 2$ &   \\
    & Fully-connected & $N \times H$ & ReLU\\
    & Fully-connected & $N \times H$ & None\\
    & Average & $H$ & \\
\hline
\multirow{2}{*}{${\rm RNN}(\bm{\alpha}_t^d)$} & Embedding $\bm{\alpha}_t^d$ & $d \times E$ &   \\
& ${\rm GRU}(\bm{\alpha}_t^d)$ last output & $2E$ &   \\
\hline
\multirow{3}{*}{MLP} & Concatenate $[{\rm DeepSets}(\psi_t), {\rm RNN}(\bm{\alpha}_t^d) ]$ & $H + 2E$ &   \\
& Fully-connected & $H$ & ReLU\\
& Output layer & $2$ &  None\\
\hline
\end{tabular}
\caption{RNN configuration.}
\label{table:RNN}
\end{table}

\section{Policy and value networks over time}
\begin{figure}[!h]
\centering
\includegraphics[width=\textwidth]{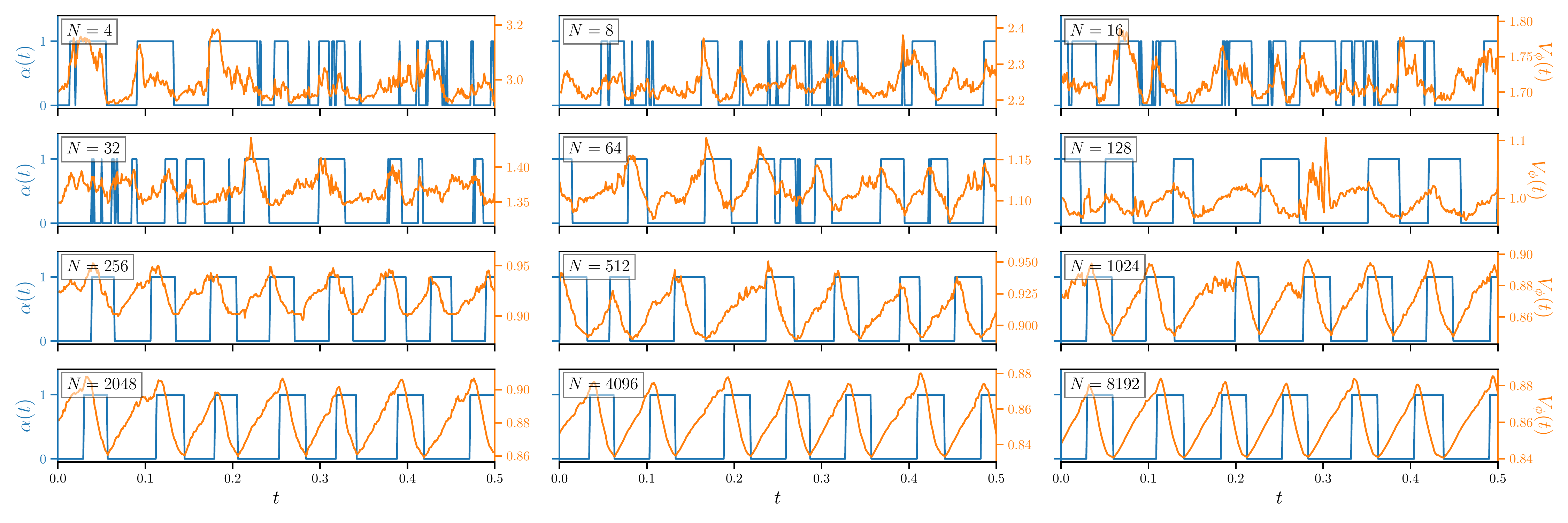}
\vskip -0.1in
\caption{Policy and value networks over time. The blue and orange lines denote $\alpha(t)$ and $V_{\phi}(t)$, respectively, as a function of time $t$ for $N=2^2, 2^3, \dots, 2^{13}$.}
\vskip -0.1in
\label{figS:time-policy}
\end{figure}

Figure~\ref{figS:time-policy} shows that with increasing $N$, the deterministic control $\alpha(t)$ of the trained DeepSets policy as a function of time $t$ converges with that of the periodic switching policy.

\section{Sawtooth potential}
\begin{figure}[!h]
\centering
\includegraphics[width=\textwidth]{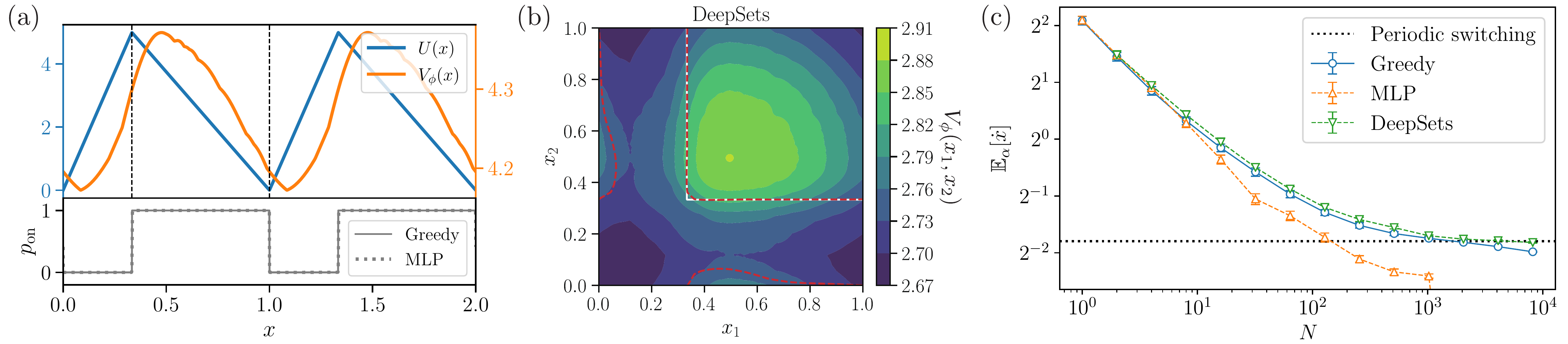}
\vskip -0.1in
\caption{Training results for the sawtooth potential. (a) $N=1$ case. Top: Sawtooth potential $U$~\eqref{eqs:sawtooth} and trained value network $V_\phi$ as a function of position $x$ are denoted by blue and orange lines, respectively. Bottom: The solid line denotes the probability of switching on the potential ($p_{\rm on}$) as a function of $x$ for the greedy policy. The dotted line represents $p_{\rm on}$ of the trained MLP policy. (b) $N=2$ case. The decision boundaries of the trained DeepSets policy and the greedy policy are shown with the red dashed and solid white contours, respectively. (c) The current $\mathbb{E}_{\alpha}[\dot{x}]$ as function of $N$ for each policy.}
\vskip -0.1in
\label{figS:sawtooth}
\end{figure}

We also test the deep RL method with the sawtooth potential~\eqref{eqs:sawtooth} with $L=1$ and $U_0=5$. The training results are shown in Fig.~\ref{figS:sawtooth}.
\begin{align}
U(x) = \left\{
    \begin{array}{lll}
        \displaystyle \frac{3U_0}{L} x & \mbox{for} & 0 \leq x \leq \frac{L}{3}  \\ && \\ 
        \displaystyle U_0 - \frac{3U_0}{2L} \left(x-\frac{L}{3}\right) & \mbox{for} & \frac{L}{3} < x < L, \\
    \end{array}
\right. {\rm and~} U(x+L)=U(x).
\label{eqs:sawtooth}
\end{align}
%